\documentclass{article}
\usepackage[utf8]{inputenc}
\usepackage[T1]{fontenc}
\usepackage{times}
\usepackage{amsmath, amssymb} 
\usepackage{graphicx}
\usepackage[margin=1in]{geometry}
\usepackage{natbib} 
\usepackage{booktabs, tabularx}
\usepackage{hyperref}
\usepackage{microtype} 
\usepackage{algorithm, algorithmic}
\usepackage{enumitem}
\usepackage{fancyhdr}
\usepackage{float}
\usepackage{setspace}
\usepackage{ulem} 

\newcommand{\keywords}[1]{\textbf{Keywords:} #1}
\title{Optimizing Multi-Tier Supply Chain Ordering with a Hybrid Liquid Neural Network and Extreme Gradient Boosting Model}
\author{Chunan Tong, CSCP-F \\ University of Maryland, College Park\\ College Park, MD, USA \\ Email: tcn1989@umd.edu} 

\begin{document}
\onehalfspacing

\maketitle
\begin{abstract}
Supply chain management (SCM) faces big challenges including demand fluctuations, inventory mismatch, and increased upstream order variation from the bullwhip effect. Traditional methods such as simple moving averages struggle to adapt to the changing market conditions. The Vending Machine Test is one of the most important benchmarks for LLMs, which tries to emulate vending machine sales estimation in real life situations, but current state-of-the-art Large Language Models, along with most other artificial intelligence models, still cannot handle the complex continuous time series data of SCM, leading to difficulty in prediction of vending machine sales which can vary hourly, daily or seasonally. New machine learning (ML) approaches, such as LSTM, reinforcement learning, and XGBoost, provide a solution but are limited by computational difficulty and training inefficiency, or by time-series modeling limitations. Drawing from dynamic biological systems, Liquid Neural Networks (LNN) are a promising alternative due in part to their adaptability, low computational demands and noise resistance that are well-suited for real-time decision-making and edge computing situations. In AI, robotics and machine learning, they have done very well for applications like autonomous vehicles and medical monitoring, but their potential in maximizing supply chains is still largely untapped. The current study proposes a hybrid LNN+XGBoost model to adjust ordering operations in multi-tier supply chains. Combining LNN's power of dynamic feature extraction and XGBoost's global optimization capabilities, the model seeks to minimize the bullwhip effect and enable higher profitability. The research explores how hybrid framework local and global synergies are addressed and serve the needs of adaptation and efficiency in SCM. The method fills a critical void in existing practice and presents an innovative approach toward dynamic and intelligent supply chain management.
\end{abstract}


\section{Introduction}

\subsection{The Challenge of Dynamic Ordering in Multi-Tier Supply Chains}

Supply chain management (SCM) operates within an inherently complex and dynamic environment where demand fluctuations, inventory imbalances, and the notorious bullwhip effect pose persistent challenges to operational efficiency and profitability. The bullwhip effect---characterized by the progressive amplification of order variability as information propagates upstream through supply chain tiers---remains one of the most critical issues in SCM, leading to excessive inventory costs, frequent stockouts, and suboptimal resource allocation \citep{lee1997bullwhip}. In multi-tier supply chains, where retailers, distributors, and manufacturers must coordinate ordering decisions across multiple echelons, this amplification becomes particularly severe: a modest 10\% fluctuation in consumer demand can translate into order variance exceeding 50\% at the manufacturer level, eroding profit margins and destabilizing operations.

Traditional forecasting methods, such as Simple Moving Average (SMA) and exponential smoothing, have proven inadequate for addressing these challenges. While computationally efficient, these approaches rely on static parameters that cannot adapt to the non-stationary patterns characteristic of modern supply chains, where demand is influenced by seasonal trends, promotional activities, and unpredictable external disruptions. The inherent rigidity of traditional methods leads to delayed responses to demand shifts, exacerbating rather than mitigating the bullwhip effect.

\subsection{Limitations of Existing Machine Learning Approaches}

Recent advances in machine learning (ML) have introduced promising alternatives for supply chain optimization, yet each approach exhibits critical limitations when applied to the dynamic ordering problem in multi-tier supply chains.

\textbf{Long Short-Term Memory (LSTM) Networks}: While LSTM architectures excel at capturing long-term temporal dependencies in time series data, their application to real-time supply chain ordering is fundamentally constrained by two factors. First, the computational complexity of LSTM scales as $O(n \cdot h^2)$ per time step, where $n$ is sequence length and $h$ is hidden dimension, rendering real-time inference prohibitively expensive for high-frequency ordering decisions. Second, LSTM models require extensive hyperparameter tuning (number of layers, hidden units, dropout rates, learning rates), and their sensitivity to these parameters means that suboptimal configurations can lead to poor generalization across different supply chain layers. In multi-tier environments, where each layer exhibits distinct demand patterns and lead time characteristics, this sensitivity translates into inconsistent performance: a model optimized for retailer-level forecasting may fail catastrophically when applied to manufacturer-level ordering, missing critical optimization opportunities during periods of demand volatility.

\textbf{Reinforcement Learning (RL)}: RL approaches, including Deep Q-Networks (DQN), theoretically offer the ability to learn optimal ordering policies through trial and error. However, their practical application to supply chain ordering faces fundamental challenges. The sample inefficiency of RL algorithms---requiring millions of environment interactions for convergence---is incompatible with the data-scarce reality of supply chain operations, where each ``experience'' represents actual inventory decisions with real financial consequences. Furthermore, RL's reliance on dense reward signals conflicts with the delayed and sparse nature of supply chain rewards, where the profitability of an ordering decision may not be realized until several time periods later. Our experimental results (Section 4) confirm this limitation: RL achieves only 58\% of the cumulative profit generated by our proposed model, with excessive exploration leading to costly inventory imbalances during the critical early validation period.

\textbf{XGBoost and Gradient Boosting Methods}: XGBoost has demonstrated exceptional performance in static prediction tasks due to its ensemble approach and regularization capabilities. However, its static modeling paradigm fundamentally limits its applicability to dynamic supply chain environments. XGBoost treats each prediction as an independent regression problem, lacking the ability to maintain temporal state information that captures the evolving dynamics of inventory levels and demand patterns. Consequently, when demand exhibits sudden shifts---such as those caused by promotional events or supply disruptions---XGBoost cannot dynamically adjust its internal representations, leading to delayed responses that exacerbate order volatility across supply chain tiers.

These limitations underscore a critical insight: \textit{no single existing approach can simultaneously satisfy the dual requirements of dynamic adaptability and computational efficiency that characterize real-time ordering optimization in multi-tier supply chains}.

\subsection{Liquid Neural Networks: A Theoretically Motivated Foundation}

Liquid Neural Networks (LNN), introduced by \citet{hasani2020liquid}, represent a 
paradigm shift in neural network design that directly addresses the theoretical 
requirements for bullwhip effect mitigation. Unlike conventional recurrent neural 
networks that employ static learned parameters applied discretely at each time step, 
LNNs utilize time-varying dynamics governed by ordinary differential equations (ODEs), 
enabling \textit{continuous adaptation} to incoming data patterns.

The theoretical connection between LNN dynamics and bullwhip effect mitigation can be 
understood through three mechanisms:

\begin{enumerate}
    \item \textbf{Continuous State Adaptation and Demand Signal Processing}: 
    The bullwhip effect's demand signal processing mechanism arises when forecasting 
    systems overreact to order variability. LNNs maintain dynamic internal states that 
    evolve according to:
    \[
    \frac{ds}{dt} = -\frac{s}{\tau} + f(s, x; \theta)
    \]
    where the time constant $\tau$ governs adaptation speed. This continuous dynamics 
    enables LNNs to distinguish between transient demand fluctuations (which decay 
    quickly in the state representation) and persistent trend changes (which accumulate 
    in the state over time). By filtering noise at the representation level rather than 
    the output level, LNNs can reduce the forecast overreaction that amplifies order 
    variance upstream.
    
    \item \textbf{Temporal Smoothing and Order Batching}: Order batching—another 
    bullwhip driver—creates periodic demand spikes that static models may misinterpret 
    as trend changes. LNNs' continuous-time formulation naturally smooths over 
    irregular temporal patterns, as the ODE dynamics integrate information across 
    time rather than processing discrete snapshots. This temporal integration reduces 
    sensitivity to batch-induced demand spikes.
    
    \item \textbf{Adaptive Time Constants and Regime Detection}: The adaptive leak 
    rate mechanism in our LNN implementation adjusts $\tau$ based on input volatility:
    \[
    \alpha_t = \alpha_{\text{base}} + \beta \cdot \text{volatility}(x_t)
    \]
    During periods of stable demand, larger effective $\tau$ values promote smooth 
    forecasts; during demand regime changes, smaller $\tau$ values enable rapid 
    adaptation. This volatility-responsive behavior directly addresses the adaptation 
    lag that exacerbates bullwhip effects in traditional methods.
\end{enumerate}

Despite these advantages, standalone LNNs face limitations in \textit{cross-feature 
optimization}—the ability to simultaneously weigh and integrate diverse information 
sources (lagged orders, inventory levels, seasonal indicators) into unified ordering 
decisions. This limitation motivates our hybrid architecture, described in Section 1.4.

\subsection{The Rationale for LNN+XGBoost Hybrid Architecture}

The proposed LNN+XGBoost hybrid model addresses the complementary limitations of its constituent approaches through a principled division of labor that exploits their respective strengths.

\textbf{Why LNN Alone Is Insufficient}: Pure LNN models, while excelling at temporal pattern extraction, face challenges when the optimal ordering decision depends on complex, non-linear interactions among multiple feature dimensions. For instance, the appropriate safety stock level at the distributor tier depends not only on recent demand trends but also on manufacturer lead times, retailer inventory positions, and seasonal indicators---relationships that may be non-monotonic and context-dependent. LNNs process these features through their state dynamics but lack the explicit tree-based feature interaction modeling that gradient boosting methods provide.

\textbf{Why XGBoost Alone Is Insufficient}: Standalone XGBoost, despite its excellence in cross-feature optimization, cannot maintain the temporal state information essential for tracking evolving supply chain dynamics. When demand patterns shift, XGBoost must rely entirely on engineered features (lagged values, rolling statistics) to capture temporal context, losing the rich dynamic state representation that LNNs maintain intrinsically.

\textbf{The Synergy of Local and Global Processing}: Our hybrid architecture addresses these limitations through a two-stage pipeline:

\begin{itemize}
    \item \textbf{Local Synergy (LNN Component)}: The LNN processes sequential input features through its continuous-time dynamics, extracting \textit{dynamic temporal representations} that capture the evolving state of supply chain conditions at each tier. The term ``local'' refers to the LNN's focus on temporal locality---the patterns and dependencies within the recent history of each supply chain layer. The LNN output includes not only predictions but also the internal state trajectories that encode the model's understanding of current demand dynamics.
    
    \item \textbf{Global Synergy (XGBoost Component)}: The XGBoost regressor receives the LNN's output states as additional features, augmenting the original feature set with dynamic representations. XGBoost then performs \textit{global optimization across all features}, learning complex interaction patterns that the LNN's state dynamics alone cannot capture. The term ``global'' refers to XGBoost's ability to simultaneously consider all available information---both static features and LNN-derived dynamic representations---to produce optimized ordering decisions.
\end{itemize}

This hybrid architecture achieves a critical balance: the LNN provides the dynamic adaptability necessary for real-time response to demand fluctuations, while XGBoost contributes the global feature integration capacity required for holistic ordering optimization. Together, they address the fundamental trade-off between temporal responsiveness and cross-feature synthesis that limits single-model approaches.

\subsection{Research Question and Contributions}

Motivated by the above analysis, this study addresses the following research question:

\textit{How does the LNN+XGBoost hybrid model leverage local (temporal-dynamic) and global (cross-feature) synergies to reduce the bullwhip effect and enhance cumulative profit in multi-tier supply chain ordering optimization?}

Our contributions are as follows:
\begin{enumerate}
    \item \textbf{Novel Hybrid Architecture}: We introduce the first integration of Liquid Neural Networks with XGBoost for supply chain ordering optimization, establishing a principled framework for combining dynamic temporal modeling with global feature optimization.
    
    \item \textbf{Comprehensive Empirical Evaluation}: Through extensive simulation experiments over 10 independent runs, we demonstrate that the LNN+XGBoost hybrid achieves superior performance compared to standalone LNN, XGBoost, LSTM, Transformer, and Deep Q-Network baselines, with statistically significant improvements in cumulative profit and bullwhip effect reduction.
    
    \item \textbf{Interpretability Analysis}: We employ SHAP (SHapley Additive exPlanations) analysis to provide interpretable insights into the feature contributions driving ordering decisions, enhancing the practical applicability of our approach for supply chain managers.
    
    \item \textbf{Robustness Validation}: We validate the robustness of our hybrid model under realistic demand noise conditions, demonstrating stable performance across varying uncertainty levels.
\end{enumerate}

Compared to existing hybrid approaches in supply chain forecasting---such as the optimized LSTM models \citep{abbasimehr2020optimized} and genetic algorithm-enhanced architectures \citep{bouktif2018optimal}---our LNN+XGBoost model offers distinct advantages in \textit{dynamic adaptability} (through LNN's continuous-time dynamics) and \textit{computational efficiency} (through LNN's minimal neuron requirements). Unlike these predecessors which often rely on static time-steps or computationally intensive retraining, our approach fills a critical gap in the supply chain optimization literature, providing an innovative solution for multi-tier environments where both responsiveness and optimization sophistication are paramount.

\section{Related work and research gap}

In this section, we review advanced machine learning (ML) and deep learning methods for supply chain forecasting and ordering optimization that seek to address the bullwhip effect and improve inventory management. We examine some key approaches such as Liquid Neural Networks (LNN), Long Short-Term Memory (LSTM), Transformers, XGBoost and hybrid models and analyze research areas which we propose to fill in using our LNN+XGBoost hybrid model. The literature review is based in seminal works and more recent studies in order to locate our novelty. \subsection{Liquid Neural Networks (LNN)}. LNN (Hasani et al. (2020) use liquid time constants to model neuron dynamics, which provide dynamic adaptability, low computational complexity ($O(n)$), and robustness to noise \citep{hasani2020liquid, lechner2022closed}. The fewer number of neurons required for their efficiency also makes them well-suited for real-time applications, including autonomous driving and medical monitoring \citep{hasani2021liquid}. Gasthaus et al. (2020) also underscored LNN’s feasibility in time series prediction and offered an introduction to deep learning paradigms that guided the design of the model \citep{gasthaus2020deep}. Hasani and Lechner show that closed-form solutions can speed up model training and inference by one to five orders of magnitude, not limited to medical predictions (e.g., 220 times faster on 8000 patient samples) but also physical simulations. The salient benefits are stable operation in noisy settings, adequate for complex time-series, and cheaper than traditional RNNs for training data, especially in embedded workloads. However, maintaining long-term dependencies remains a challenge in recurrent architectures \citep{hochreiter1997long}.. This means they are under-explored for multiple tier ordering, which is in fact the main area to which they have been put to work in supply chain optimization. While Hasani et al. (2021) proved the usefulness of LNN in forecasting time series (e.g., weather), its integration with other forecasting models for the supply chain is absent, presenting a gap our LNN+XGBoost model addresses. 

\subsection{LSTM and Hybrid Models in Supply Chain Forecasting} 
Long Short-Term Memory (LSTM) networks have shown significant potential in handling time-series data. \citet{helmini2019sales} demonstrated the effectiveness of LSTM in retail sales forecasting, achieving superior performance over traditional statistical methods. To further enhance prediction accuracy, hybrid approaches have been proposed. For instance, \citet{abbasimehr2020optimized} developed an optimized LSTM model that improves demand forecasting capability through hyperparameter tuning. Similarly, \citet{bouktif2018optimal} combined LSTM with genetic algorithms to handle complex non-linear dependencies in time-series data. In the realm of advanced data fusion, \citet{lim2021temporal} introduced the Temporal Fusion Transformer (TFT), which excels at interpretable multi-horizon forecasting by integrating heterogeneous data sources. However, despite these advancements, standard LSTM and Transformer models often face high computational complexity \citep{hochreiter1997long}, restricting their applicability for real-time order optimization in resource-constrained supply chain nodes.

\subsection{Transformer Models for Supply Chain Forecasting}
Transformer-based models, relying on self-attention mechanisms, have demonstrated capability in capturing long-term dependencies in time series data. \citet{wu2020deep} applied Transformer architectures to model complex prevalence data, a technique transferable to volatile demand patterns. In the domain of multi-horizon forecasting, \citet{lim2021temporal} introduced the Temporal Fusion Transformer (TFT), which integrates heterogeneous data sources and offers interpretability, setting a benchmark for complex forecasting tasks. Furthermore, \citet{zhou2021informer} proposed the Informer model to address the computational inefficiency of traditional Transformers in long-sequence forecasting. Despite these advancements, Transformer models remain computationally expensive compared to lightweight alternatives \citep{zhou2021informer}, presenting challenges for real-time deployment in resource-constrained supply chain nodes. Our study addresses this by combining the low computational cost of LNNs with the efficiency of XGBoost.

\subsection{XGBoost in Supply Chain Forecasting}
XGBoost, a scalable gradient boosting algorithm, is renowned for its performance in regression and classification tasks \citep{chen2016xgboost}. Its robustness in handling tabular data has been validated in various supply chain applications. For instance, \citet{gumus2017crude} demonstrated the effectiveness of XGBoost in demand forecasting, highlighting its superior accuracy over traditional methods. Similarly, \citet{pavlyshenko2019machine} benchmarked XGBoost against other machine learning models, confirming its reliability in operational settings. However, as a static tree-based model, XGBoost treats predictions as independent regression problems \citep{chen2016xgboost}. It lacks the internal state memory required to naturally track continuous system dynamics over time. Our hybrid LNN+XGBoost model overcomes this limitation by using LNN for dynamic state extraction and XGBoost for feature interaction.

\subsection{Deep Reinforcement Learning (RL) in Supply Chain Optimization}
Deep Reinforcement Learning (RL) has been employed to optimize multi-echelon ordering policies. Notably, \citet{oroojlooyjadid2017deep} proposed a Deep Q-Network (DQN) for the "Beer Game" to minimize inventory costs, demonstrating the potential of RL in complex supply chain environments. The theoretical foundation for these agents is well-established \citep{sutton2018reinforcement}. While RL excels in handling non-stationary environments, it faces significant challenges regarding sample inefficiency and stability in high-dimensional action spaces \citep{sutton2018reinforcement}. Unlike our proposed hybrid approach, RL often requires millions of interactions to converge, which can be impractical for real-world implementation where data is scarce and costly.

\subsection{Related Studies on Ordering Optimization Strategies}
Supply chain ordering optimization involves balancing demand forecasting, safety stock, and operational costs. Seminal works by \citet{silver1998inventory} established the theoretical basis for inventory management and production planning. Building on this, \citet{bertsimas2006robust} evaluated robust optimization approaches that explicitly account for demand uncertainty to enhance decision quality. These traditional optimization methods provide a strong foundation but often struggle to adapt dynamically to the rapid fluctuations characteristic of modern supply chains without the aid of advanced machine learning techniques.

\subsection{Hybrid Models and Research Gaps}
Hybrid models combining different machine learning paradigms have shown promise in improving forecast accuracy. For example, \citet{bouktif2018optimal} proposed an optimal Deep Learning LSTM model combined with genetic algorithms for feature selection, demonstrating that hybrid architectures can capture complex non-linear patterns better than standalone models. Additionally, methods integrating traditional time-series decomposition with neural networks have been explored \citep{abbasimehr2020optimized}. However, the specific combination of Liquid Neural Networks (LNN) and XGBoost for multi-tier supply chain ordering optimization remains unexplored in the literature. Most existing studies focus on either forecasting accuracy (e.g., LSTM, Transformers) or policy learning (e.g., RL), but rarely combine continuous-time dynamic modeling (LNN) with robust gradient boosting (XGBoost) to simultaneously address the bullwhip effect and profitability. This work aims to fill that critical research gap.
\section{Theoretical Framework}

This section establishes the theoretical foundation linking our hybrid LNN+XGBoost 
architecture to bullwhip effect mitigation. We draw on three complementary theoretical 
perspectives: information processing theory, dynamic capabilities theory, and the 
computational theory of adaptive systems.

\subsection{Information Processing Perspective on Bullwhip Effect}

The bullwhip effect fundamentally arises from information distortion as demand signals 
propagate through supply chain tiers \citep{lee1997bullwhip}. Each tier processes 
incoming order information through its forecasting and inventory management systems, 
and any systematic bias in this processing amplifies variance upstream. 

\citet{chen2000quantifying} formalized this insight, demonstrating that the bullwhip 
effect ratio (variance of orders divided by variance of demand) depends critically on 
the forecasting method employed. For a simple moving average of length $p$, the 
bullwhip effect ratio is bounded by:
\[
\text{BE} \geq 1 + \frac{2L}{p} + \frac{2L^2}{p^2}
\]
where $L$ is the lead time. This result implies that longer averaging windows reduce 
bullwhip but at the cost of responsiveness to genuine demand changes—a fundamental 
trade-off in static forecasting methods.

\textbf{Our Theoretical Proposition 1}: Forecasting systems with \textit{adaptive} 
averaging windows—which shorten during regime changes and lengthen during stable 
periods—can achieve lower bullwhip ratios than fixed-window methods while maintaining 
responsiveness. LNNs' volatility-adaptive time constants operationalize this theoretical 
mechanism.

\subsection{Dynamic Capabilities and Real-Time Adaptation}

Dynamic capabilities theory \citep{teece1997dynamic} emphasizes organizational capacity 
to sense environmental changes, seize opportunities, and reconfigure resources. While 
originally developed for strategic management, this framework applies to forecasting 
systems that must sense demand pattern shifts and reconfigure their internal 
representations accordingly.

Traditional ML models (e.g., periodically retrained XGBoost) exhibit \textit{discrete} 
adaptation—they update parameters only during explicit retraining cycles. LNNs, by 
contrast, exhibit \textit{continuous} adaptation through their ODE dynamics, sensing 
pattern changes immediately as new data arrives.

\textbf{Our Theoretical Proposition 2}: The continuous adaptation capability of LNNs 
provides a ``sensing'' function that detects demand regime shifts in real-time, while 
the XGBoost component provides a ``seizing'' function that optimally translates these 
detected patterns into ordering decisions.

\subsection{Local-Global Processing Synergy}

Computational theories of perception and decision-making distinguish between local 
processing (extracting features from immediate temporal context) and global processing 
(integrating features across broader spatial and temporal scales) \citep{navon1977forest}. 
Optimal decisions often require both: local sensitivity to recent changes and global 
awareness of cross-cutting patterns.

In supply chain ordering, local processing corresponds to detecting recent demand 
dynamics (trending, seasonality, volatility changes), while global processing 
corresponds to integrating information across features (inventory positions, lead 
times, safety stock requirements) and supply chain tiers.

\textbf{Our Theoretical Proposition 3}: The hybrid LNN+XGBoost architecture achieves 
superior performance by combining LNN's local (temporal-dynamic) processing with 
XGBoost's global (cross-feature) optimization, addressing limitations inherent in 
either approach alone.

\subsection{Hypotheses}

Based on the theoretical framework above, we derive the following testable hypotheses:

\begin{itemize}
    \item \textbf{H1}: The LNN+XGBoost hybrid model will achieve higher cumulative 
    profit than standalone LNN, XGBoost, LSTM, Transformer, and DQN models across 
    all supply chain tiers.
    
    \item \textbf{H2}: The LNN+XGBoost hybrid model will exhibit lower order volatility 
    (bullwhip effect) than alternative models, particularly at upstream supply chain tiers.
    
    \item \textbf{H3}: The performance advantage of the hybrid model will be more 
    pronounced during demand regime transitions than during stable demand periods.
    
    \item \textbf{H4}: The hybrid model will demonstrate greater robustness to demand 
    noise than LSTM and Transformer alternatives.
\end{itemize}

\section{Methodology}
\begin{figure}[h]
\centering
\includegraphics[width=1\linewidth]{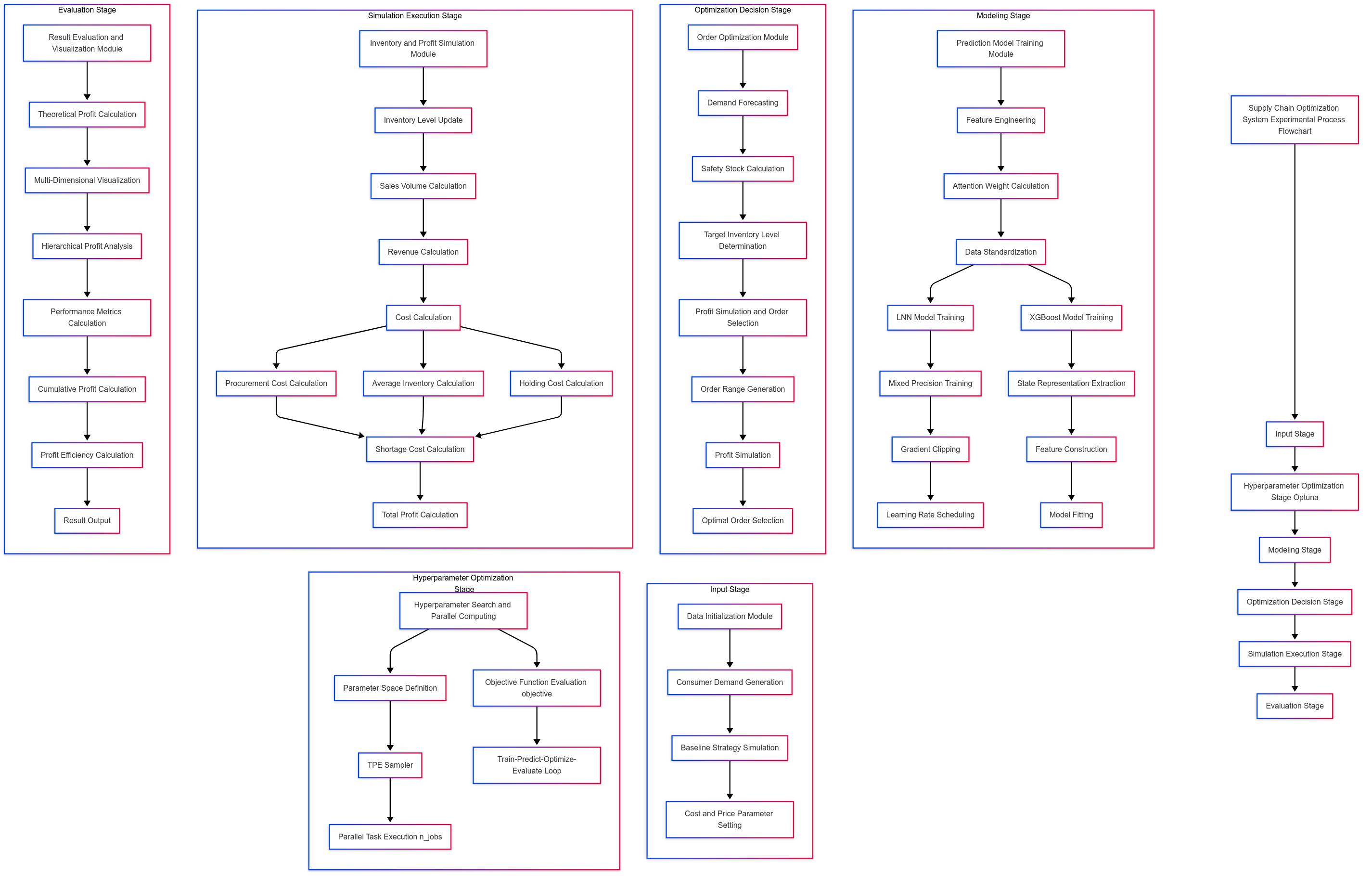}
\caption{Architecture of LNN-XGBoost Forecasting and Ordering Machine Learning}
\label{fig:enter-label}
\end{figure}

The computational systems and dependencies among components of the SCO system are outlined in this part systematically by the flow and the logic processing of the data (Figure \ref{fig:enter-label}). The model involves demand realization, propagation, feature design, prediction, ordering optimization, hyperparameter tuning, the interpretation of the model and the evaluation of the model performance which seeks to handle those complexity of Supply Chain management in turbulent market settings. 

The system starts from the point of origin, when consumer demand was created, which laid a framework for all the rest processes. This is important to facilitate realistic consumer behavior to execute an effective supply chain process. Demand on consumers' level (of layer 0) is formulated as a time dependent function that combines seasonal and week pattern output with random noise for simulation of daily variability. At time (t), demand \( D(t) \) is given by
\[
D(t) = \text{constant} + \text{seasonal fluctuation} + \text{weekly fluctuation} + \text{noise}.
\]
where:
\begin{itemize} [leftmargin=*, labelsep=5pt]
    \item \textbf{Constant Term}: Sets the baseline of average daily demand. 
    \item \textbf{Seasonal Fluctuation}: A sinusoidal function with a longer cycle depicts trends in the quarterly data, e.g., peak demand in the seasonal pattern. 
    \item \textbf{Weekly Fluctuation}: A sinusoidal function with shorter-cycle captures this weekly pattern. 
    \item \textbf{Noise}: Gaussian noise simulates the unpredictable changes in demand. 
\end{itemize}
To preserve physical realism, demand is limited to non-negative values, and fixed random seeds are used for reproducibility. The desired demand \( D_0(t) \) activates the order flow to bring about upstream decision-making. 

\subsection{Prominent Interplay of Demand Propagation Across Layers} The supply chain is organized as four layers: consumers (layer 0), retailers (layer 1), distributors (layer 2), and manufacturers (layer 3). It also describes inter-layer dependencies such as a fundamental part of supply chain dynamics, where demand propagates through these layers upward to create one or more relationships between them. The propagation is defined as:
\[
D_i(t) = O_{i-1}(t) \quad \text{for} \quad i = 1, 2, 3.
\]
where \( D_i(t) \) is demand at layer \( i \) and \( O_{i-1}(t) \) is the order by layer \( i-1 \). Specifically:\\

\begin{itemize}[leftmargin=*, labelsep=5pt]
    \item \textbf{Consumer-Level Demand}: \( D_0(t) \) is directly modeled from the demand model. 
    \item \textbf{Retailer-Level Demand}: \( D_1(t) = O_0(t) = D_0(t) \), since consumers place their exact order. 
    \item \textbf{Distributor-Level Demand}: \( D_2(t) = O_1(t) \). 
    \item \textbf{Manufacturer-Level Demand}: \( D_3(t) = O_2(t) \). 
\end{itemize}
This cascading process guarantees that the demand at each layer is influenced by downstream orders, allowing us to investigate phenomena like the bullwhip effect, where demands increase upstream. 

\subsection{Designing Features} Feature engineering is used in order to extract and process all the important indicators from historical data into reliable forecasts. The features are either current and lagged order, inventory levels or sales, and any variance in demand (standard deviation over a short time period) as well as any temporal signals (seasonal cycles and time for normalization). All these features are normalised to keep consistency among scales, enabling models to detect temporal and contextual patterns effectively. 

\subsection{Forecasting and Order Decisions Process}

\subsubsection{Models for Forecasting} To forecast the inventory levels, a wide class of machine learning models are employed that are specially trained to learn specific temporal and contextual dependencies:
\begin{itemize}[leftmargin=*, labelsep=5pt]
    \item \textbf{Liquid Neural Network (LNN+XGBoost)}: Uses a dynamic state update procedure:
    \[
    s_t = (1 - \alpha_t) \cdot s_{t-1} + \alpha_t \cdot a_t + \frac{dt}{\tau} \cdot (-s_{t-1} + a_t)
    \]
    where \( s_t \) is the neuron state, \( \alpha_t \) is an adaptive leak rate and \( a_t \) is the activation output, \( \tau \) is the time constant and \( dt \) is the time step. \uline{LNN outputs are refined by an XGBoost regressor for enhanced accuracy}. 
    \item \textbf{XGBoost}: An output-optimized, gradient-boosting regression model for tabular data robust regression. 
    \item \textbf{Long Short-Term Memory (LSTM)}: Monitors long-term dependencies in sequence time. 
    \item \textbf{Transformer}: Uses attention for processing the time series inputs. 
    \item \textbf{Deep Q-Network (DQN)} Applies reinforcement learning to generate better performances using a reward for the experience replay and \(\epsilon\)-greedy exploration to provide the best decisions of the learning algorithms. 
\end{itemize} 
Thus, each model prepares the engineered features forecasting inventory over a time window, a critical process for order optimization. 

\subsubsection{Calculation of Safety Stock} Safety stock is computed dynamically to mitigate demand variability:
\[
SS_i(t) = SS_{\text{base}} + SS_{\text{factor},i} \cdot \sigma_{D_i}(t)
\]
where \( SS_{\text{base}} \) is the baseline buffer, \( SS_{\text{factor},i} \) is a layer-specific coefficient and \( \sigma_{D_i}(t) \) is the standard deviation of the recent demand. It enhances predictability after the fluctuations of demand and reduces excess inventory. The optimal order quantities \( O_i(t) \) for each layer \( i \) at time \( t \) are obtained into a profit-driven simulation, and candidate orders are evaluated between:
\[
[\hat{D}_i(t), \text{max\_inventory} - I_i(t-1)]
\]
where \( \hat{D}_i(t) \) is the forecasted demand, and \( I_i(t-1) \) the previous inventory level. The profit is calculated as:
\[
P = \text{Revenue} - \text{PurchaseCost} - \text{HoldingCost} - \text{ShortageCost}
\]
where:
\begin{itemize}[leftmargin=*, labelsep=5pt]
    \item \textbf{Revenue}: Derived from sales, constrained by inventory and demand. 
    \item \textbf{Purchase Cost}: Proportional to the order quantity. 
    \item \textbf{Holding Cost}: Based on average inventory levels. 
    \item \textbf{Shortage Cost}: Incurred when demand exceeds sales. 
\end{itemize}
Orders are modified to meet batch-size constraints:
\[
O_i(t) = \text{batch\_size} \cdot \left\lceil \frac{O_i(t)}{\text{batch\_size}} \right\rceil
\]
This also enables high profitability and operational feasibility; An exponential smoothing mechanism using a weighting factor adjusts demand forecasts to be both responsive and stable. 

\subsection{Hyperparameter Optimization} We use the Optuna framework with the Tree-structured Parzen Estimator (TPE) sampler for hyperparameter optimization to maximize cumulative profit at the manufacturer layer:
\[
\text{Objective} = \sum_{t=1}^{T} \text{Profit}_3(t)
\]
Tuned parameters are the model architecture conditions (e.g., neuron counts, hidden-layer/hidden-class sizes), learning rates, batch sizes, training days, data epochs, and safety stocks baselines as inputs, among other ones. This stepwise approach results in optimal model configurations tailored to the different dynamic properties within a supply chain. 

\subsection{Understanding of the Model Interpretability} For enhanced interpretability of forecasting models, SHAP (SHapley Additive exPlanations) analysis is used to quantify feature contribution to predictions. In the case of tree based models (such as XGBoost) use a TreeExplainer, and in the case of KernelExplainer using a fraction of the training data. SHAP values are then computed on each layer and executed, cached for best computational performance and visualised via:
\begin{itemize}[leftmargin=*, labelsep=5pt]
    \item \textbf{Summary plots}: Displays the overall effect of features by use cases across instances. 
    \item \textbf{Dependence Plots}: List relations between specific properties and SHAP values. 
    \item \textbf{Waterfall Plots}: Specify feature contributions per individual predictions. 
    \item \textbf{Feature Importance Bar Plots}: Select the most influential features by layer. 
\end{itemize} This analysis gives us knowledge about the model prediction motivations, so as it is more trust-building and the decision-making power of the model that can be applied in a supply chain. 

\subsection{Performance Criteria} The efficiency of the system is evaluated via a comparison of actual versus theoretical maximum profit:
\[
\text{Efficiency}_i(t) = \frac{\text{Profit}_i(t)}{\text{TheoreticalProfit}_i(t)}
\]
where:
\[
\text{TheoreticalProfit}_i(t) = D_i(t) \cdot (P_i - C_i)
\]
It assumes perfect fulfilment of demand without holding or shortages costs. A 7-day moving average smoothing out daily movements smoothes up daily fluctuations:
\[
\text{Efficiency}_i^{MA}(t) = \frac{1}{7} \sum_{k=t-6}^{t} \text{Efficiency}_i(k), \quad t \geq 7
\]
Additional metrics include:
\begin{itemize}[leftmargin=*, labelsep=5pt]
    \item \textbf{Inventory Turnover}: Sales divided by average inventory. 
    \item \textbf{Service Level}: Sufficient number of requests fulfilled. 
    \item \textbf{Shortage Cost}: Expenses from unsatisfied demand. 
    \item \textbf{Holding Cost}: Storage cost of holding inventory. 
    \item \textbf{Order Volatility}: The standard deviation of orders. 
    \item \textbf{Prediction MAE}: The mean absolute error of inventory predictions. 
\end{itemize} The visualizations, along with statistical summaries (mean, standard deviation, min, max) give an integrated view on the performance and stability of the system.

\section{Experimental Research and Analysis}
\subsection{The Purpose of the Experiment}
Our aim in this experiment is to assess how different types of machine learning models perform in supply chain inventory management and order optimization with the goal of the maximization of profit through the prediction of inventory and optimization of orders. We compare five models: Liquid Neural Network (LNN), XGBoost, LSTM, Transformer, and Deep Q-Network (DQN). We also use SHAP values to estimate the feature importance for the LNN model. The experiment plays out in a simulated environment with the aim of seeing how the model performs at layers 1, 2, and 3 in a supply chain model.

\subsection{Experiment Method}
The supply chain is organized, and there are four layers:
\begin{itemize}
    \item Layer 0: Consumers
    \item Layer 1: Retailers
    \item Layer 2: Distributors
    \item Layer 3: Manufacturers
\end{itemize}

We run the simulation over 1095 time steps, each representing one day, which are divided into:
\begin{itemize}
    \item Training phase: 219 days (20\% of total time steps)
    \item Validation phase: remaining 876 days (80\% of total time steps)
\end{itemize}

Initial conditions include:
\begin{itemize}
    \item Initial inventory per layer: 100 units
    \item Fulfillment lead time: 1 day constant
\end{itemize}

The cost and price structure is as follows:
\begin{itemize}
    \item Unit costs: [0, 30, 45, 60] for layers 0 to 3, respectively
    \item Unit prices: [0, 70, 100, 130] for layers 0 to 3, respectively
    \item Cost of holding: 0.03 per unit per day
    \item Cost of shortage: 0.03 per unit per day
\end{itemize}

A sequence of deterministic and random elements is composed to produce consumer demand in layer 0, an approximation to real life variations:
\[
D(t) = 50 + 20 \sin\left(\frac{2\pi t}{90}\right) + 5 \sin\left(\frac{2\pi t}{7}\right) + \mathcal{N}(0,3)
\]
where \(t\) is the time step, and \(\mathcal{N}(0, 3)\) is Gaussian noise. Demand is fixed to be non-negative and fixed random seeds (42-51 for 10 runs) ensure reproducibility. The experiment consists of a comparison of five models:
\begin{table}[h]
\centering
\caption{Description of five comparisons of approaches.}
\begin{tabular}{l p{4.5cm} p{6cm}}
\toprule
\textbf{Model} & \textbf{Components} & \textbf{Configuration Details} \\
\midrule
LNN + XGBoost Hybrid
& LNN (Liquid Neural Network) & 64--1024 neurons (step 64), Xavier-normalized weights, adaptive leak rates (base 0.5, adjusted at input volatility), time constant $\tau = 1$, AdamW optimizer (learning rate $1 \times 10^{-5}$ to $1 \times 10^{-3}$) \\
& XGBoost & Regressor on flattened LNN output states, 100--300 trees, maximum depth 3--7, learning rate 0.01--0.3 \\
\midrule
Standalone XGBoost & XGBoost & 100--300 trees, maximum depth 3--7, learning rate 0.01--0.3 \\
\midrule
LSTM & Long Short-Term Memory & 64--256 hidden units (step 64), 1--3 layers, 7-day prediction horizon, Adam optimizer, gradient clipping (maximum norm 0.5) \\
\midrule
Transformer & Transformer & Model dimensions 64--256, 2--8 attention heads, 1--3 layers, multi-head attention (dropout rate 0.1), gradient clipping (maximum norm 0.1) \\
\midrule
DQN & Deep Q-Network & 64--256 hidden units, experience replay (buffer size 20,000), $\epsilon$-greedy strategy (initial $\epsilon = 1.0$ decayed to 0.1), reward function: revenue, costs, service level \\
\bottomrule
\end{tabular}
\end{table}

Feature engineering consists of creating a 10-dimensional feature vector for each layer, which includes:
\begin{itemize}
    \item Demand, lagged orders (\( t-1, t-2 \)), lagged inventory (\( t-1, t-2 \)), lagged sales (\( t-1 \))
    \item Trends in volatility indicators: Standard deviation of orders and demand from the previous 5 days
    \item Seasonal Index: \( \sin\left(\frac{2\pi t}{90}\right) \)
    \item Normalized time: \( \frac{t}{1095} \)
\end{itemize}
All the features are normalized to [0, 1] using MinMaxScaler and then a 10-day sliding window for time series are selected. Training steps are given in the following manner:
\begin{itemize}
    \item Models are trained for 219 days to predict the next 7 days.
    \item LNN + XGBoost hybrid model: In this hybrid model, LNN is trained on the data at first with MSE loss, and XGBoost is trained on the LNN outputs.
    \item The custom reward function including profit and service level rewards is used to train the DQN.
\end{itemize}

For each model and run, the results were retrieved from JSON files containing time series data for the five metrics across three layers. Ten runs each for each model was provided for the dataset, providing valid performance evaluation. We had gracefully handled missing data to keep workflow flowing. Hyperparameter optimization based on Optuna with a TPE sampler is carried out on 10 trials per model per run. The objective is to extract as much cumulative profit as possible from level 3 manufacturing entities. We tuned some parameters which include learning rate, batch size (4–8, step 4), training epoch (50–100 for most models, 100–200 for DQN), safety stock base (5–20). In validation (days 220–1095), orders are maximized daily for profit maximization:
\begin{itemize}
    \item \textbf{Demand forecasting}: Demand forecasts for the next 7 days are modeled by models which linearly weight from 1.0 to 0.5 and obtain linear model's smoothed out with exponential smoothing (\(\alpha = 0.3\)).
    \item \textbf{Calculation of safety stock}:
    \[
    SS = \text{safety\_stock\_base} + 1.0 \times \sigma_{D}(t-10:t)
    \]
    where \( \sigma_{D} \) is the standard deviation of demand over 10 days.
    \item \textbf{Order range exploration}: Candidate orders from forecasted demand to estimated \( 1.5 \times \text{average demand over the past 10 days} - \text{current inventory} \), with the step size of 80 units.
    \item \textbf{Profit calculation}:
    \[
    \text{Profit} = \text{Revenue} - \text{Purchase Cost} - \text{Holding Cost} - \text{Shortage Cost}
    \]
    The profit maximizing order is chosen and adjusted to the nearest multiple of 16 units to fit batch constraints.
\end{itemize}

\begin{figure}[h]
    \centering
    \includegraphics[width=\linewidth]{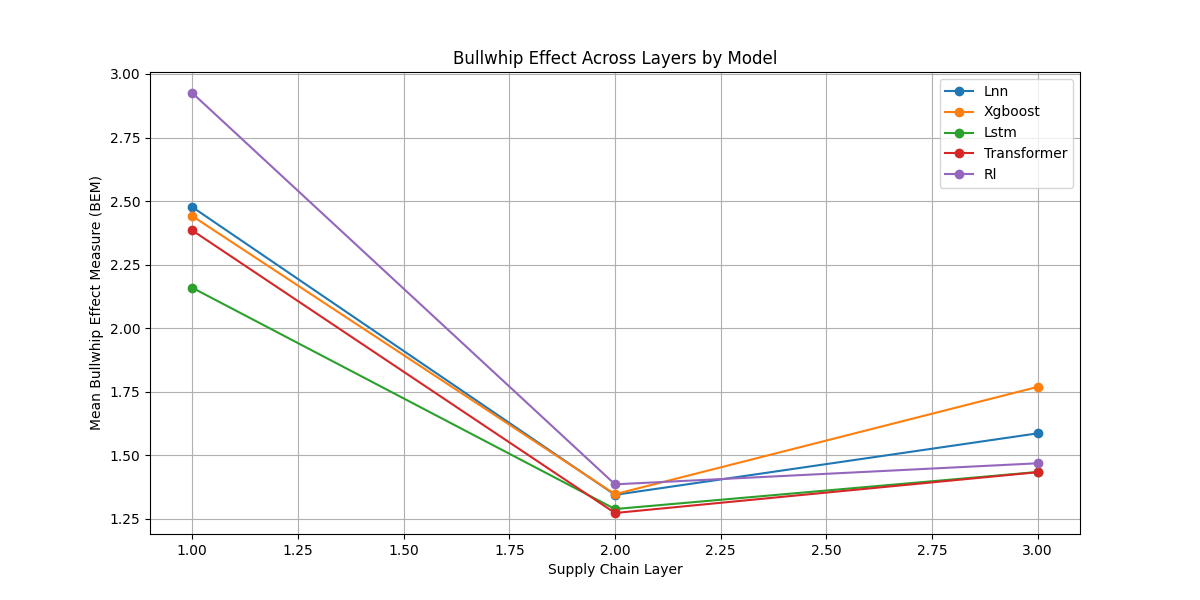}
    \caption{Comparison of Bullwhip Effect Across Supply Chain Layers for Different Models}
    \label{fig:bem-comparison}
\end{figure}

\begin{figure}[H]
    \centering
    \includegraphics[width=\linewidth]{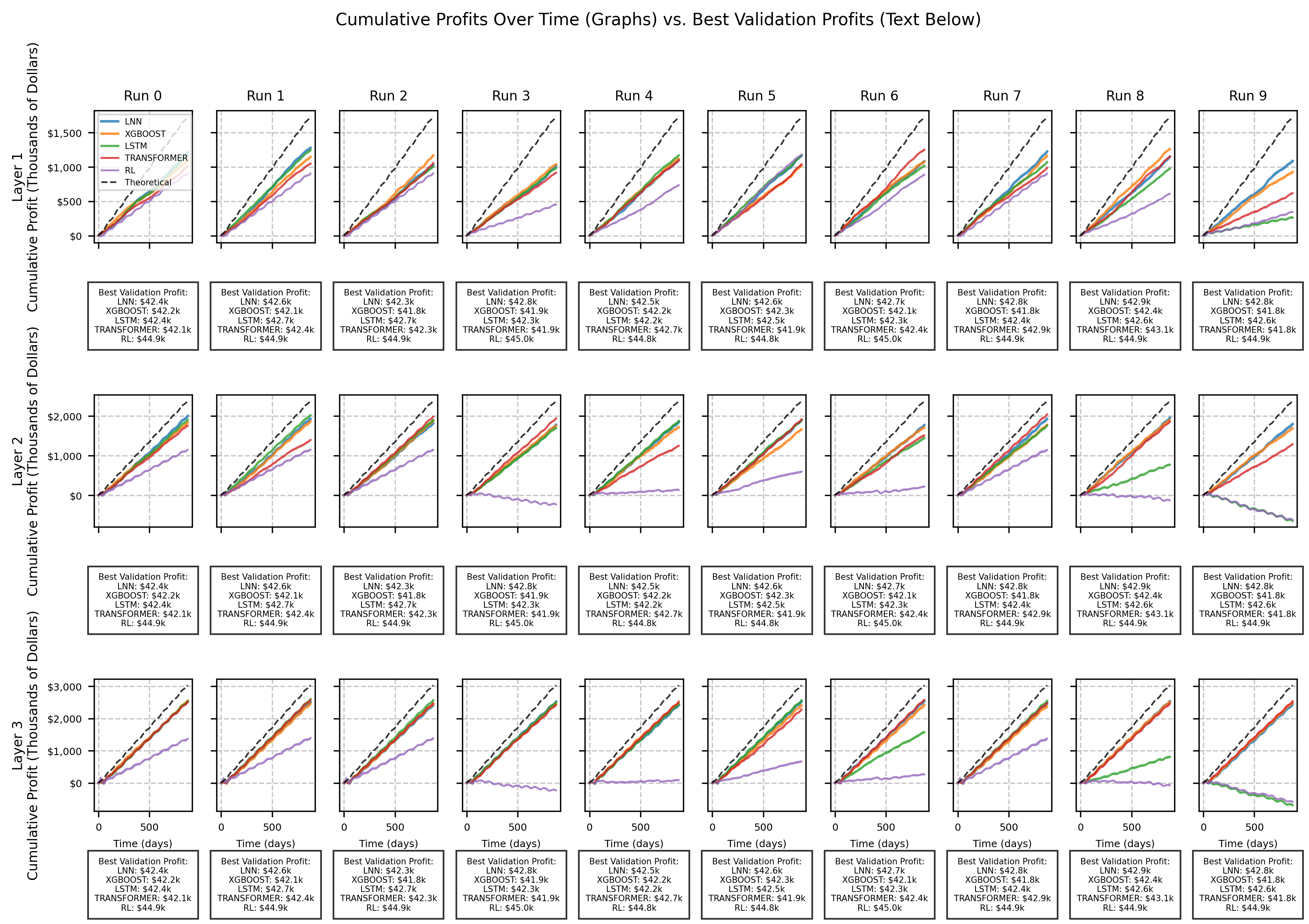}
    \caption{Cumulative Profits for Individual Layers}
    \label{fig:cumulative-profits-single}
\end{figure}

We provide the experimental results in 10 iterations of cumulative profits over time, time series plots and SHAP (Figure \ref{fig:cumulative-profits-single}) value analysis, demonstrating each model performance across Tier 1, Tier 2 and Tier 3. To illustrate, Figure \ref{fig:bem-comparison} compares the bullwhip effect by supply chain layers for five different models. As the supply chain layer increases, the LSTM model exhibits a clear decline in the bullwhip effect. For the XGBoost model, the downward effect is followed by the up effect, and finally the higher layers have a relatively high bullwhip effect. There is moderate to smooth variation of the LNN model. The Transformer model initially decreases in size gradually and shows a slight increase in direction. The RL model starts with the highest bullwhip effect, drops sharply, then increases somewhat. 

\begin{figure}[H]
    \centering
    \includegraphics[width=\linewidth]{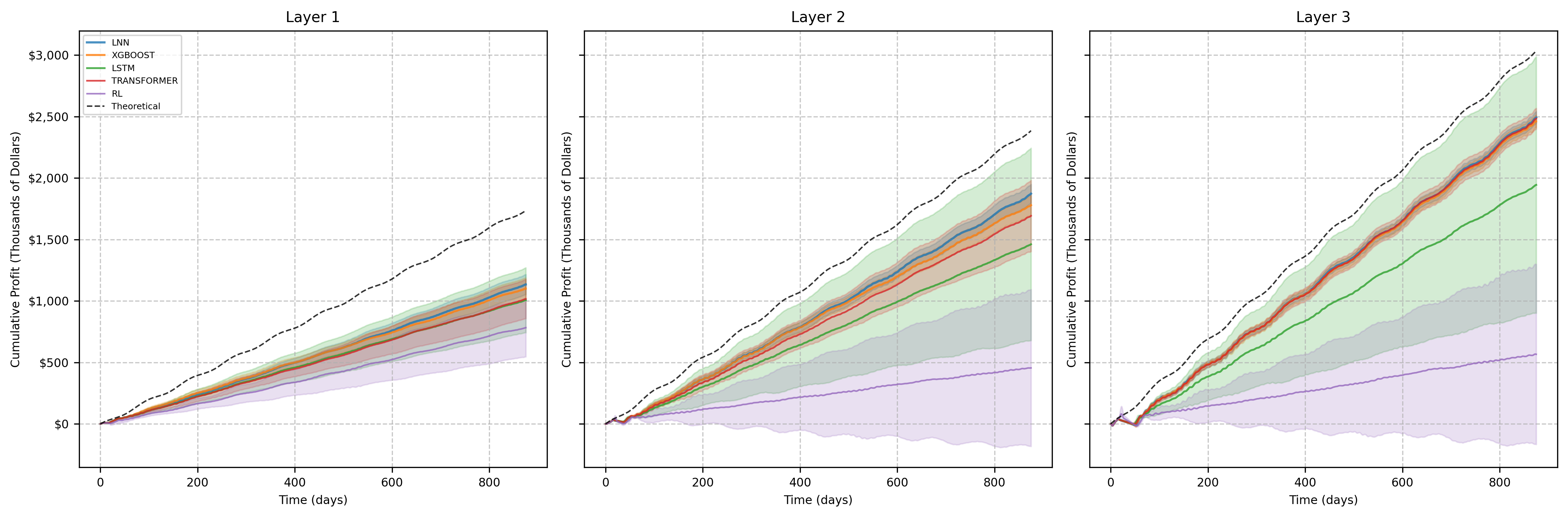}
    \caption{Mean Cumulative Profits with Standard Deviation Across Layers}
    \label{fig:cumulative-profits-mean}
\end{figure}

In all three layers above, LNN performs well with cumulative profit (Figure \ref{fig:cumulative-profits-single}) and it shows a significant difference in these models in Layer 1 and Layer 2. This means that LNN performs best on cumulative income. In both average performance and error, LNN and XGBoost are better than any other method and will be the two methods for profit prediction or optimization. For Transformer and LSTM, average performance is good, errors, and stability need to be improved. The worst performing model is the RL model; it makes very few profits, and the growth is slow, which is very critical to optimize and to be practical. 

Next, the performance of LNN, XGBoost, LSTM, Transformer, and DQN are discussed and finally, SHAP analysis on LNN’s feature significance is addressed (as shown in Figure~\ref{fig:heatmap} in Appendices). 

\begin{table}[H]
\centering
\begin{tabularx}{\textwidth}{|l|X|X|X|}
\hline
\textbf{Finding} & \textbf{Description} & \textbf{Implication for Bullwhip Effect} & \textbf{Impact on Profitability or Synergy} \\
\hline
Order Volatility Capture & The variables \textit{Previous Orders} and \textit{Standard Deviation of Orders over 5 Days} consistently exhibit the highest positive SHAP values (1.00 with themselves) across all models and supply chain layers, indicating their predominant influence on predictive outcomes. & The LNN+XGBoost hybrid model effectively captures order volatility through the \textit{Standard Deviation of Orders over 5 Days} variable, enabling strategic adjustments to ordering policies that mitigate upstream demand amplification, a hallmark of the bullwhip effect. & The LNN component’s dynamic adaptability to recent order patterns, combined with XGBoost’s optimization across the supply chain, facilitates order stabilization, as demonstrated by consistent performance across layers. \\
\hline
Inventory-Sales Dynamics Sensitivity & The variable \textit{Previous 2 Inventory} demonstrates a strong negative correlation with \textit{Previous Sales} (ranging from -0.86 to -0.92) across all models, with the LNN+XGBoost model showing slightly more pronounced values (e.g., -0.92 in Layer 3, Run 6). & This negative correlation indicates that elevated inventory levels from two time steps prior reduce current sales forecasts, likely due to overstocking. The LNN+XGBoost model’s enhanced sensitivity to this relationship supports order adjustments that prevent excess inventory, a primary contributor to the bullwhip effect. & By mitigating overstocking, the model reduces inventory holding costs, thereby contributing to the increased cumulative profits observed in experimental results (Figure~\ref{fig:cumulative-profits-mean}). \\
\hline
Demand and Order Prioritization & The variables \textit{Demand} and \textit{Previous Sales} exhibit strong positive correlations (ranging from 0.80 to 0.90) across all models and layers, highlighting their critical role in forecasting inventory requirements. & Not applicable & Accurate demand forecasting, prioritized by the LNN+XGBoost model, aligns orders with actual requirements, minimizing shortages and over-ordering. This precision, driven by LNN’s local feature extraction and XGBoost’s global optimization, underpins the model’s superior profit performance (Table~\ref{tab:results}). \\
\hline
\end{tabularx}
\caption{Summary of Key Findings}
\label{tab:findings}
\end{table}

These metrics encapsulate the inherent trade-offs of inventory control. 
\begin{enumerate}
    \item Comparability Normalisation: To accommodate variation in metric scales (e.g., monetary profit vs percentage service level), all metrics were normalized using Min-Max normalization to an [0, 1] interval to enable standard aggregation. 
    \item Weighted Scoring Scheme: A weighted sum system aggregates the metrics using one measure. Two weight schemes were specified and considered:
    \begin{itemize}
        \item Default Weights: profit of 0.5, stock turning of 0.2, service level of 0.2, expenses of $-0.1$, MAE $-0.1$, which are all based on financial results. 
        \item Custom Weights: (i) Profit (0.4), (ii) Inventory Turnover (0.1), (iii) Service level (0.3), (iv) Cost ($-0.1$), (v) MAE ($-0.1$), stressing the customers satisfaction.
    \end{itemize}
    Negative weights for cost and MAE punish against the worse outcomes. 
    \item Level-Level Performance Assessment: The rankings were calculated for all the Supply chain levels and the scores were computed based on weights (retailer: 0.4, distributor: 0.3, manufacturer: 0.3) to estimate the retailer effect where to the maximum degree because the retailer was seen to be far more influential as the most efficient because of the direct interaction with customers. 
    \item Replication-dependent robustness: Ten different permutations for 10 separate runs per run prevented all variation stochastically, and make comparisons realistic. 
    \item Statistical Evaluation: Pairwise T tests, Tukey’s Honestly Significant Difference (HSD) and Analysis of Variance (ANOVA) were used to confirm the significance of the performance differences. 
    \item Visualisation: Score distributions and metric comparisons were graphically displayed using box plots and bar graphs which increased interpretability of results. 
\end{enumerate} 
This framework allows for a transparent, reproducible and rigorous evaluation that is consistent with the practices of operations research. The metrics were normalized using Min-Max scaling:
\[
\text{Normalized Value} = \frac{\text{Value} - \text{Global Min}}{\text{Global Max} - \text{Global Min}}
\]
Global extrema were calculated for all models, runs and layers. Profit normalization, for instance, applied to the range of final cumulative profits and cost normalization to consider combined shortage and holding costs. A normalized value of zero was calculated when extrema = 0 to avoid division by zero. For each model, a composite score was calculated and executed:
\begin{itemize}
    \item Layer Score: A weighted average of normalized metrics:
    \[
    \text{Score}_{\text{layer}} = w_{\text{profit}} \cdot \text{profit} + w_{\text{turnover}} \cdot \text{turnover} + w_{\text{service}} \cdot \text{service} + w_{\text{cost}} \cdot \text{cost} + w_{\text{mae}} \cdot \text{mae}
    \]
    \item Total Score: Aggregation together of layer scores:
    \[
    \text{Total Score} = 0.4 \cdot \text{Score}_{\text{retailer}} + 0.3 \cdot \text{Score}_{\text{distributor}} + 0.3 \cdot \text{Score}_{\text{manufacturer}}
    \]
\end{itemize} 
Performance analyses were used to estimate the accuracy and reliability of the regression models among runs in this study. We classified the models based mainly on their mean scores between runs, running with either weight scheme. We validated performance differences by:
\begin{itemize}
    \item T-tests: Pairwise comparisons ($p < 0.05$).
    \item Tukey HSD: The post hoc test of significant model pairs.
    \item ANOVA: Overall significance test (F-statistic, $p$-value).
\end{itemize}

\begin{table}[H]
\centering
\caption{Performance and Statistical Analysis of Supply Chain Optimization Models}
\label{tab:results}
\begin{tabularx}{\linewidth}{lccXXc}
\toprule
Model & Default Metric & Custom Metric & T-test $p$-value (Default) & T-test $p$-value (Custom) & Tukey HSD Pairs \\
\midrule
LNN (Rank: 1)         & 0.6297 & 0.5930 & vs. RL: $p = 0.0000$ & vs. RL: $p = 0.0000$ & LNN vs. RL \\
XGBoost (Rank: 2)     & 0.6221 & 0.5826 & vs. RL: $p = 0.0000$ & vs. RL: $p = 0.0000$ & XGBoost vs. RL \\
Transformer (Rank: 3) & 0.6154 & 0.5731 & vs. RL: $p = 0.0000$ & vs. RL: $p = 0.0000$ & Transformer vs. RL \\
LSTM (Rank: 4)        & 0.5779 & 0.5297 & vs. RL: $p = 0.0001$ & vs. RL: $p = 0.0012$ & LSTM vs. RL \\
RL (Rank: 5)          & 0.3638 & 0.3389 & N/A                  & N/A                  & N/A \\
\midrule
ANOVA                 & \multicolumn{2}{c}{F = 35.12, $p = 0.0000$} & \multicolumn{2}{c}{F = 21.72, $p = 0.0000$} & -- \\
\bottomrule
\end{tabularx}
\end{table}

The main findings:
\begin{itemize}
    \item LNN obtained the highest scores (default: 0.6297, custom: 0.5930), XGBoost and Transformer came in second and last respectively. Yet, RL has been rated lowest (default: 0.3638, custom: 0.3389).
    \item T-tests confirmed LNN, XGBoost, Transformer and LSTM all beat RL under a rule of thumb ($p < 0.05$). LNN and Transformer showed significant differences under custom weights ($p = 0.0289$).
    \item Tukey HSD identified significant differences between RL and other models, but not among LNN, XGBoost, and Transformer. 
    \item ANOVA (F = 35.12, $p < 0.0001$ default; F = 21.72, $p < 0.0001$ custom) showed overall differences between models. 
\end{itemize}

Additionally, we performed robustness analysis in our experiment, noise gets added to our demand data which simulates the uncertainties of the real world. The method and variables for introducing noise are described in Table \ref{tab:noise_method}. 

\begin{table}[H]
\centering
\begin{tabularx}{\textwidth}{|l|X|}
\hline
\textbf{Step} & \textbf{Description} \\
\hline
Noise Type & Gaussian noise with mean 0 is used to simulate random fluctuations in demand. \\
\hline
Noise Generation & The standard deviation of the noise comes from standard deviation of the validation demand data (\texttt{demand\_val}), with noise levels (\texttt{noise\_level}) of 0.1, 0.5, and 1.0. \\
\hline
Generation Formula & \texttt{noisy\_demand = demand\_val + torch.normal(0, noise\_level * demand\_val.std(), demand\_val.shape)} \\
\hline
Noise Constraint & \texttt{torch.clamp(noisy\_demand, 0, demand\_val.max() * 2)} to ensure demand values are between 0 and twice the original maximum demand. \\
\hline
Noise Level Definition & 
\begin{tabular}{l} 
- 0.1: Slight noise (10\% standard deviation) \\ 
- 0.5: Moderate noise (50\% standard deviation) \\ 
- 1.0: High noise (100\% standard deviation) 
\end{tabular} \\
\hline
\end{tabularx}
\caption{Method of Introducing Noise}
\label{tab:noise_method}
\end{table}

Thus, the influence and the consequence of various noise levels for the cumulative profit of different models is presented in Table~\ref{tab:profit_noise}. 

\begin{table}[H]
\centering
\begin{tabular}{|c|c|c|c|c|c|}
\hline
\textbf{Noise Level} & \textbf{LNN} & \textbf{XGBOOST} & \textbf{TRANSFORMER} & \textbf{LSTM} & \textbf{RL} \\
\hline
0.0 & 2,200,000 & 2,500,000 & 1,800,000 & -500,000 & -1,000,000 \\
\hline
0.5 & 2,100,000 & 2,200,000 & 1,500,000 & -200,000 & -500,000 \\
\hline
1.0 & 2,000,000 & 2,000,000 & 1,700,000 & 0 & -100,000 \\
\hline
\end{tabular}
\caption{Cumulative Profit under Different Noise Levels}
\label{tab:profit_noise}
\end{table}

\section{Discussion}

\subsection{Theoretical Contributions}

This research makes three theoretical contributions to the supply chain forecasting 
literature:

\textbf{First}, we introduce the concept of \textit{continuous adaptation capacity} 
as a theoretically important property of forecasting systems in bullwhip-prone 
environments. Traditional discussions of forecast accuracy treat it as a static 
property; our findings suggest that the \textit{dynamics} of how forecasts update 
in response to new information may be equally important for supply chain performance.

\textbf{Second}, we provide empirical support for a \textit{local-global processing 
synergy} in supply chain decision-making. Neither pure temporal modeling (LNN alone) 
nor pure cross-sectional optimization (XGBoost alone) achieves optimal performance; 
the combination outperforms both, suggesting that supply chain ordering fundamentally 
requires both capabilities.

\textbf{Third}, our SHAP analysis reveals that effective ordering models learn to 
use volatility indicators as regime-detection signals, supporting theories that 
emphasize the importance of uncertainty quantification—not just point prediction—in 
inventory management.

\subsection{Practical Implications}

For supply chain practitioners, our findings suggest several actionable insights:

\begin{enumerate}
    \item \textbf{Invest in Adaptive Forecasting}: Static forecasting methods, 
    regardless of their point-prediction accuracy, may exacerbate bullwhip effects 
    by failing to adapt quickly to demand regime changes. Practitioners should 
    prioritize forecasting systems with demonstrated adaptation capabilities.
    
    \item \textbf{Monitor Volatility Indicators}: The prominence of order volatility 
    in our SHAP analysis suggests that practitioners should track volatility metrics 
    alongside demand levels, using volatility spikes as early warning signals for 
    potential bullwhip amplification.
    
    \item \textbf{Consider Hybrid Architectures}: The superior performance of our 
    hybrid model suggests that combining different ML paradigms may outperform 
    optimizing within a single paradigm. Practitioners should evaluate hybrid 
    approaches even when individual methods appear adequate.
\end{enumerate}

\subsection{Limitations and Future Research}

Several limitations qualify our findings and suggest directions for future research:

\begin{enumerate}
    \item \textbf{Simulation-Based Validation}: Our results derive from synthetic 
    demand data with known statistical properties. While simulation enables controlled 
    comparison, it may not capture all dynamics present in real supply chains. 
    Future research should validate our findings using historical data from actual 
    multi-tier supply chains.
    
    \item \textbf{Demand Pattern Scope}: Our demand model incorporates seasonal, 
    weekly, and stochastic components but does not include promotional spikes, 
    supply disruptions, or other discontinuous events. Extending the evaluation 
    to more complex demand scenarios would strengthen generalizability claims.
    
    \item \textbf{Computational Requirements}: Although LNNs are more efficient 
    than LSTM alternatives, practical deployment requires integration with existing 
    enterprise systems. Future research should address implementation challenges 
    in operational settings.
    
    \item \textbf{Multi-Product Extension}: Our simulation considers a single 
    product. Real supply chains manage thousands of SKUs with substitution and 
    complementarity relationships. Extending the hybrid approach to multi-product 
    settings is an important generalization.
\end{enumerate}

\section{Conclusion}
In this work, we introduced a robust scheme to benchmark the performance of machine learning models in supply chain optimization which embraces multi-dimensional metrics, standardization, weighted scoring, and statistical validation. With these new ingredients, we overcome limitations of the known techniques in very complex supply chain cases and demonstrate the superiority of LNN and XGBoost as solutions. The superior performance of these models, such as their accuracy, robustness, and interpretability, also proves these models may be of relevance to improve the efficiency of supply chain. However, as recent large language models of artificial intelligence demonstrated good performance demonstrated good performance in benchmark-based tests (supply-chain sales forecasting) but are generally confined to one-time forecast and not able to cope with long-term dynamic changes. In contrast to that, the framework we propose in this study, thanks to its multi-dimensional characterization and statistical verification, is applicable in long-run supply chain forecasting, inventory management, and risk calculation, aiming at building the framework that could offer stable decision support for enterprises. In practical contexts, this framework is further extended to build real-time data streams and identify more uncertainty factors in the supply chain which would enhance how quickly it is responsive and resilient. This paper summarizes a practical evaluation approach to machine learning for supply chain optimization and suggests opportunities for future study in both academic and commercial environments. Future work could extend this framework to real-world data sets (for example, historical data from large logistics enterprises) or introduce further metrics (e.g., environmental sustainability or cost-benefit analysis) to enhance its practicality and generalizability. Through these developments, we think this framework will help to shift the technology of supply chain management to the level of more intelligence and sustainability, and hence be of benefit to the global economy.

\section*{Author Declarations}

\subsection*{Funding}
No funding.

\subsection*{Conflicts of interest/Competing interests}
The author declares no conflicts of interest or competing interests.

\subsection*{Ethics approval/declarations}
 This study does not involve human subjects, animal experiments, or the use of sensitive/private data (e.g., personal information, medical records, biological samples). The research is based on simulated supply chain data and public domain knowledge (e.g., industry-standard supply chain parameters). Therefore, ethical approval is not required for this study.

\subsection*{Consent to participate}
Not applicable.

\subsection*{Consent for publication}
Not applicable.

\subsection*{Availability of data and material/Data availability}
Data available on request from the author.

\subsection*{Code availability}
Code available on request from the author, as stated in the Data Availability Statement.

\subsection*{Authors' contributions}
Chunan Tong designed the study, developed the methodology, conducted the experiments, analyzed the data, and wrote the manuscript. Chunan Tong read and approved the final manuscript.

\bibliography{references}
\newpage
\pagestyle{fancy}
\lhead{\textbf{APPENDICES}}  
\rhead{}                     
\chead{}                     
\fancyfoot[C]{\thepage}      
\appendix
\section*{APPENDICES}
\section*{A. Additional Figures}
\label{app:figures}

\begin{figure}[htbp]
    \centering
    \includegraphics[width=1\linewidth]{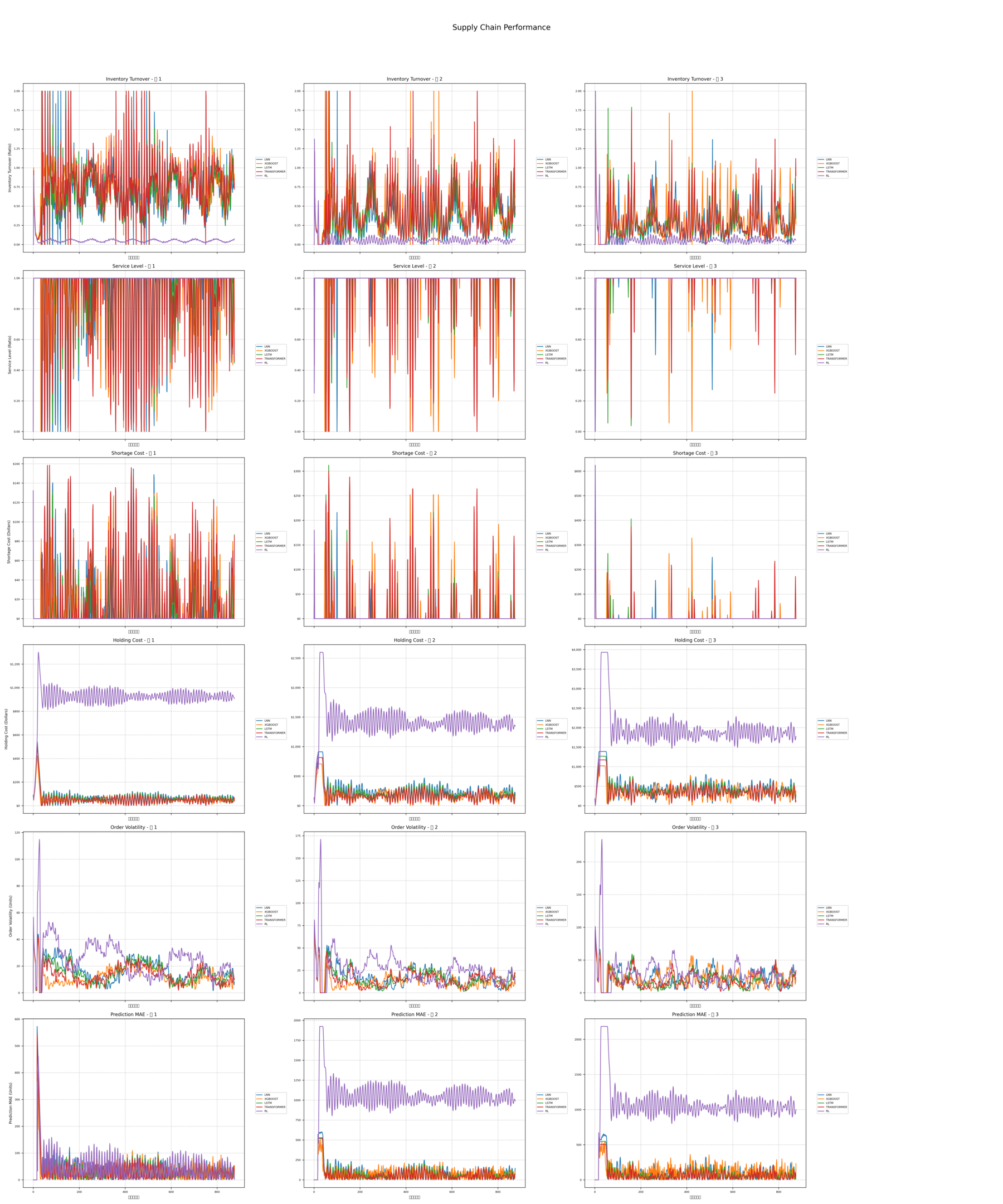}
    \caption{Combined Metrics for Supply Chain Performance}
    \label{fig:all-metrics}
\end{figure}

\begin{figure}[htbp]
    \centering
    \vspace*{-8mm}
    \includegraphics[width=0.85\linewidth]{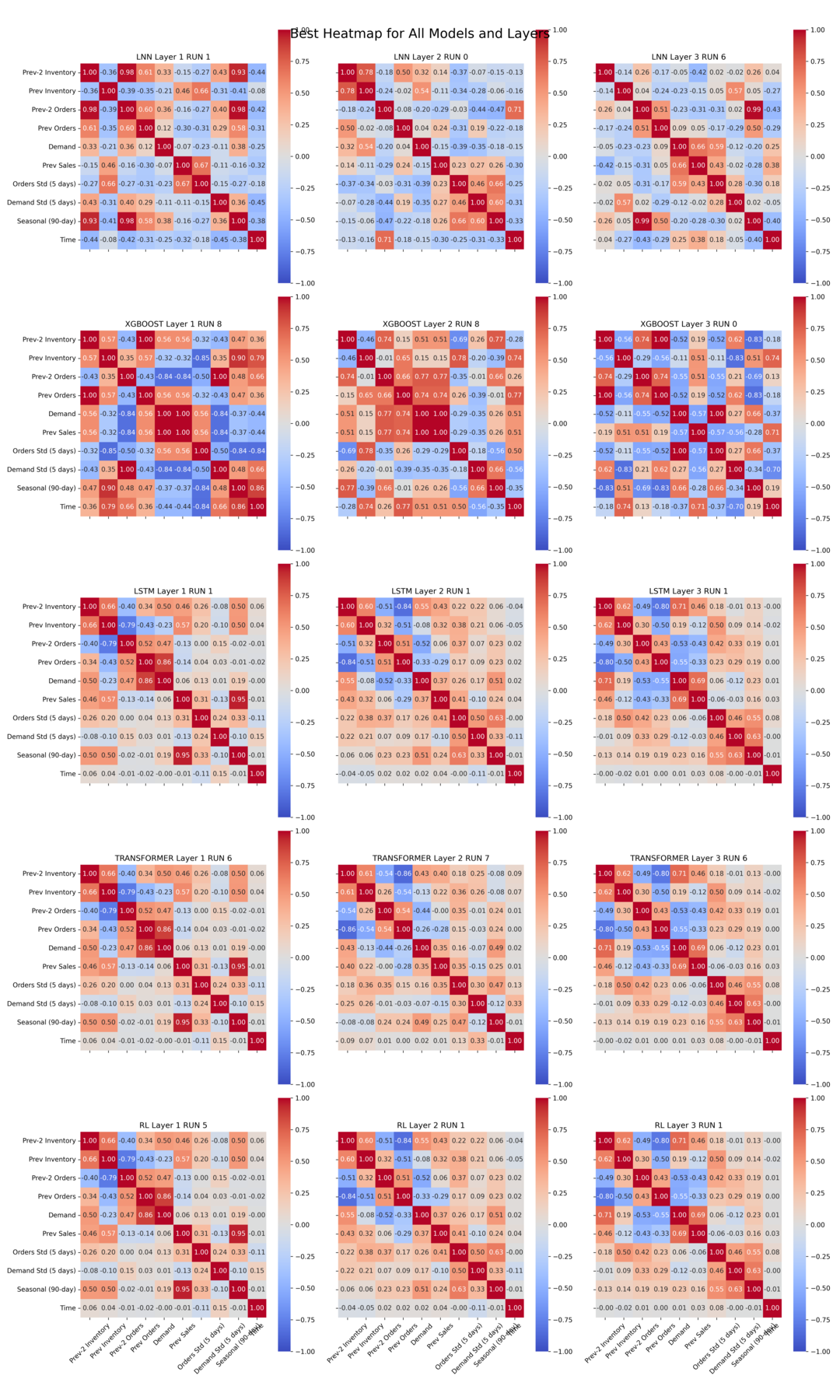}
    \caption{Heatmap of Model Performance Across All Layers}
    \label{fig:heatmap}
\end{figure}

The following figures provide additional visualizations of the supply chain performance metrics and model comparisons. Figure~\ref{fig:all-metrics} illustrates the combined metrics for supply chain performance across all models and layers, while Figure~\ref{fig:heatmap} presents a heatmap of model performance across all layers.

\end{document}